\documentclass[nohyperref]{article}

\usepackage{microtype}
\usepackage{graphicx}
\usepackage{subfigure}
\usepackage{booktabs} 

\usepackage{hyperref}

\usepackage[accepted]{icml2023}

\usepackage{amsmath}
\usepackage{amssymb}
\usepackage{mathtools}
\usepackage{amsthm}

\usepackage[capitalize,noabbrev]{cleveref}

\theoremstyle{plain}

\theoremstyle{definition}

\theoremstyle{remark}

\usepackage[textsize=tiny]{todonotes}

\icmltitlerunning{The Role of Linguistic Priors in Measuring Compositional Generalization of Vision-Language Models}

\begin{document}

\twocolumn[
\icmltitle{The Role of Linguistic Priors in Measuring \\
           Compositional Generalization of Vision-Language Models}



\icmlsetsymbol{aws}{$\dagger$}

\begin{icmlauthorlist}
\icmlauthor{Chenwei Wu}{duke,aws}
\icmlauthor{Li Erran Li}{amazon}
\icmlauthor{Stefano Ermon}{amazon,stanford}
\icmlauthor{Patrick Haffner}{amazon}
\icmlauthor{Rong Ge}{duke}
\icmlauthor{Zaiwei Zhang}{amazon}
\end{icmlauthorlist}

\icmlaffiliation{duke}{Department of Computer Science, Duke University}
\icmlaffiliation{amazon}{AWS AI, Amazon}
\icmlaffiliation{stanford}{Department of Computer Science, Stanford University}

\icmlcorrespondingauthor{Chenwei Wu}{chenwei.wu592@duke.edu}

\icmlkeywords{Machine Learning, ICML}

\vskip 0.3in
]



\printAffiliationsAndNotice{\internaws}  

\begin{abstract}
Compositionality is a common property in many modalities including natural languages and images, but the compositional generalization of multi-modal models is not well-understood. In this paper, we identify two sources of visual-linguistic compositionality: linguistic priors and the interplay between images and texts. We show that current attempts to improve compositional generalization rely on linguistic priors rather than on information in the image. We also propose a new metric for compositionality without such linguistic priors. 

\end{abstract}

\section{Introduction}
\label{sec:intro}
Compositional generalization is the ability to combine known concepts in a novel way. It exists in multiple modalities, and is believed to be a key ingredient to human intelligence. For example, after knowing the meanings of "a person wearing a shirt" and "a cat playing with a dog", one should be able to understand "a person playing with a cat". Similarly, after seeing a wood chair and a green apple, one could imagine a green chair. Human are known to have good compositional generalization ability. However, it is not clear whether neural networks have these capabilities, especially in a multi-modal setting.

On the one hand, text-to-image generative models, including DALL$\cdot$E 2 \cite{ramesh2022hierarchical}, Parti \cite{yu2022scaling}, and Imagen \cite{saharia2022photorealistic}, are good at generating novel combinations of known concepts, e.g., an avocado chair or a snake made of corn. On the other hand, previous research have shown that many multi-modal networks lack compositional generalization \cite{thrush2022winoground}. This seems contradictory to our impression.

Moreover, a multi-modal network obtains information from multiple modalities. It can utilize information from a single modality, and it can also benefit from combining information from multiple modalities. As for compositional generalization, it is not clear whether this ability comes from a single modality or multiple ones. For instance, when generating images from text prompts, Parti is hard to overcome strong linguistic and visual priors \cite{yu2022scaling}.

All this raises many questions. Where does the compositional generalization of multi-modal models come from? How to measure multi-modal compositional generalization and resolve the contradiction mentioned above? 

To answer these questions, we analyze the multi-modal compositional generalization in a more fine-grained way. We discovered the capability of language models on vision-language compositional generalization benchmarks and found that language models can beat most vision-language models without looking at the images, demonstrating a potential bias of these benchmarks. 

We then decompose the source of compositional generalization into two parts: the uni-modal part and the multi-modal part. We illustrate the contribution of these two parts for commonly-used models. We also propose a new metric ``hard test accuracy'' to quantify the contribution from linguistic priors into the overall compositional generalization ability of vision-language models.

\section{Related Works}
\label{sec:related_works}
Compositional generalization is frequently studied in natural languages and is also discussed in other domains including images \cite{shi2021retrieval}, videos \cite{grunde2021agqa}, and programming languages \cite{gan2022measuring}. It can be measured by various benchmarks about different aspects: For natural languages, people proposed SCAN \cite{lake2018generalization}, COGS \cite{kim2020cogs}, CommonGen \cite{lin2019commongen}, etc. For multi-modal domains including vision-language, it can be measured by different tasks, e.g., visual question answering \cite{bitton2021automatic, bogin2021covr, hudson2019gqa}, verb understanding \cite{hendricks2021probing}, counting \cite{parcalabescu2020seeing}, image-text retrieval \cite{shekhar2017foil}, and word ordering \cite{thrush2022winoground}. People have also proposed different metrics and fine-grained aspects \cite{andreas2019measuring, hupkes2020compositionality, keysers2019measuring} with benchmarks to measure compositional generalization ability.

Using these benchmarks, people have done various analysis of compositional generalization. For instance, \cite{hessel2021effective, lasri2022word, papadimitriou2022classifying} showed that BERT is generally not sensitive to word ordering, and \citet{parcalabescu2021valse, yuksekgonul2022and} found that vision-language models do not perform well at word order, relationship, or counting. 

People also have studied the interaction among modalities for multi-modal models, e.g., language bias can solve ``FOIL it!'' benchmark \citep{madhyastha2018defoiling}, the ``unimodal collapse'' phenomenon \citep{parcalabescu2021valse}, and the asymmetry of information flow between image and text modalities \citep{frank2021vision}.

\section{Preliminaries}
\label{sec:prelim}
\subsection{ARO Datasets: A benchmark for multi-modal compositional generalization}

ARO dataset is proposed by \citet{yuksekgonul2022and} to measure the compositional generalization of visual-language models. This dataset consists of 4 parts: Visual Genome (VG) Attribution, VG Relation, COCO Ordering, and Flickr Ordering. Each data point in this dataset contains an image with multiple captions (2 captions for VG tasks and 5 for Ordering tasks). One of these captions is the true caption corresponding to the image, and the other ones are corrupted captions that were obtained by changing the attribution, relation, or word ordering of the true caption. Given the image and the captions, the model is asked to pick the true caption among all candidates. Examples of the data in the ARO dataset are shown in Table \ref{tab:ARO-example}, and the details of this dataset can be found in \citet{yuksekgonul2022and}.

\subsection{Performance of Vision-language Models on ARO Datasets}

In \citet{yuksekgonul2022and}, the authors computed the prediction accuracy of some most popular vision-language models. Given a pair of image and text, the model computes a score between $0$ (or $-1$) and $1$ reflecting the alignment between the image and the text, i.e., the degree to which the text matches the image. Perfectly-aligned image and text have a score of 1. 
Similar to \citet{yuksekgonul2022and}, we use CLIP \cite{radford2021learning}, BLIP \cite{li2022blip}, and FLAVA \cite{singh2022flava} to replicate their results, as shown in the top half of Table \ref{tab:LM-ARO-acc}. BLIP and FLAVA have multiple heads that could be used to match images with texts, and we compute the accuracy of all these heads.

\begin{table}[htbp]
\caption{Example captions from ARO datasets. VG\_A/R for VG Attribution/Relation, CC/FL\_O for COCO/Flickr Ordering}
\label{tab:ARO-example}
\vskip 0.15in
\begin{center}
\begin{small}
\begin{sc}
\begin{tabular}{ll}
\toprule
Dataset & Caption Examples \\
\midrule
VG\_A & \textbf{T}: the wood floor and the black bag\\
      & \textbf{F}: the black floor and the wood bag\\
\midrule
VG\_R & \textbf{T}: the plate is on the table\\
      & \textbf{F}: the table is on the plate\\
\midrule
CC\_O & \textbf{T}: a fire hydrant on a city street\\
      & \textbf{F}: on a city a fire hydrant street\\
\midrule
FL\_O & \textbf{T}: group gathered to go snowmobiling\\
      & \textbf{F}: group go snowmobiling gathered to\\
\bottomrule
\end{tabular}
\end{sc}
\end{small}
\end{center}
\vskip -0.1in
\end{table}

As noted in \citet{yuksekgonul2022and}, the compositional generalization performance of commonly-used vision-language models are usually better than chance, but far from satisfying. This indicates the lack of compositional generalization ability for these models.

\section{Pure Language Models Perform Well on ARO Datasets}
\label{sec:LM_performance}
We may see from Table \ref{tab:ARO-example} that we can easily pick the true caption by only looking at these captions alone because the true captions usually are more likely to happen. For instance, it is very unlikely that a table is on a plate or a bag is made of wood. Besides, previous works showed that large language models are very powerful in various tasks \cite{brown2020language}. Therefore, by utilizing the knowledge only from the text side, it is possible to perform pretty well on these datasets. To validate our hypothesis, we would like to compute the prediction accuracy of language models on these ARO datasets.

\subsection{Prediction Method}
\label{subsec:LM}

\textbf{Predicting true caption with pure language modeling:} Given multiple captions of the same length, we compute their perplexity and predict the caption with the smallest perplexity as the true caption. Concretely, given an input sentence $x=(x_1, x_2, \cdots, x_n)$ that contains $n$ tokens, its perplexity computed by a language model $LM$ is defined as
\begin{equation}
\label{eqn:def-pp}
PP_{LM}(x) := -\frac{1}{n-1}\sum_{i=2}^{n}\log\Pr\nolimits_{LM}\left(x_i|x_1\cdots x_{i-1}\right).
\end{equation}
Here $\Pr_{LM}\left(x_i|x_1\cdots x_{i-1}\right)$ is the predicted probability of the $i$-th token being $x_i$ given by the language model $LM$ conditioning on the previous sequence $x_1\cdots x_{i-1}$. 

\textbf{Language Models Selection:} Since the number of parameters in the text encoder of vision-language models is usually at the scale of 100M or above, we choose the smallest version of GPT-2 \cite{radford2019language}, which has about 117M parameters, to do a fair comparison. To further investigate the compositional generalization of language models, we also use larger language models with the number of parameters ranging from 1B to 6B. These models include GPT-2-XL \cite{radford2019language}, OPT-1.3B \cite{zhang2022opt}, and GPT-J-6B \cite{gpt-j}. To reduce the variance in the prediction, we use an unweighted average version of GPT-2-XL, OPT-1.3B, and GPT-J-6B, i.e., for each caption, we use these three language models to compute the perplexities and take the average, and select the caption with the lowest average perplexity. We call this predictor ``AVG\_LM'', and this predictor will be used in later sections to build our new metric.

\textbf{Performance of Language Models on ARO Datasets:} The prediction accuracies of the language models are shown in the bottom half of Table \ref{tab:LM-ARO-acc}.

\begin{table}[htbp]
\caption{Prediction Accuracies on ARO datasets, VG\_A/R for VG Attribution/Relation, CC/FL\_O for COCO/Flickr Ordering. ITC/ITM/CONTR are ITC/ITM/contrastive heads of models. Language Model Details in Section \ref{subsec:LM}.}
\label{tab:LM-ARO-acc}
\vskip 0.15in
\begin{center}
\begin{small}
\begin{sc}
\begin{tabular}{lcccc}
\toprule
 & CC\_O & Fl\_O & VG\_A & VG\_R \\
\midrule
CLIP & 47.73 & 59.12 & 61.35 & 59.79\\
CLIP\_FT & 31.18 & 40.96 & 62.41 & 60.01\\
BLIP ITC & 13.21 & 17.60 & 80.00 & 41.59\\
BLIP ITM & 20.83 & 25.02 & \textbf{88.56} & 53.35\\
FLAVA Contr & 7.03 & 18.66 & 60.37 & 29.13 \\
FLAVA ITM & 4.81 & 13.20 & 68.83 & 23.71 \\
\midrule
GPT-2 & 95.28 & 96.26 & 76.57 & 85.07\\
\midrule
GPT-2 XL & 95.79 & 96.68 & 80.64 & 85.07\\
OPT 1.3B & \textbf{98.00} & \textbf{98.32} & 81.60 & 84.30\\
GPT-J 6B & 94.71 & 95.82 & 82.66 & 85.36\\
\midrule
avg\_LM & 97.33 & 97.70 & 84.48 & \textbf{85.67}\\
\bottomrule
\end{tabular}
\end{sc}
\end{small}
\end{center}
\vskip -0.1in
\end{table}

From Table \ref{tab:LM-ARO-acc}, we notice that for the Ordering tasks, GPT-2 is almost perfect in selecting the correct caption among all 5 candidates. This is intuitive because the false captions are obtained by permuting the words of the true caption in a fairly random way and usually destroys the sentence structure. In VG Relation, although the accuracy of GPT-2 is not close to perfect, it still outperforms the vision-language models. In VG Attribution, GPT-2 is slightly worse than BLIP, but are significantly better than the other models. Therefore, with roughly the same number of parameters, GPT-2 can achieve better or on-par performance than vision-language models. If we increase the number of parameters of the language model, the performance can be further increased.

To summarize, in most cases, it is possible to beat the vision-language models in compositional generalization by simply looking at the texts. Therefore, it is natural to ask: Does the compositional generalization ability of these vision-language models mostly come from the text side? Answering this question requires a fine-grained analysis, which we show in the next section.

\section{Fine-grained Analysis of Multi-modal Compositional Generalization}
\label{sec:NegCLIP}
The compositional generalization ability of multi-modal models consists of two parts: uni-modal compositional generalization and its multi-modal counterpart. In the case of ARO datasets, since the image is fixed for each datum, the only two possible sources of information are the text alone and the interplay between the image and the text. To investigate the contribution of these two sources to the model performances, we analyze the prediction accuracy of models conditioned on the properties of caption candidates.

\textbf{Testbed Selection:} In \citet{yuksekgonul2022and}, the authors proposed a new method to fine-tune CLIP on COCO. For each image in the COCO training set, they generate negative captions by randomly swapping two content words or phrases with the same part of speech as in the true caption.\footnote{They also did hard negative mining for images, which we do not include in this paper. However, our model trained with only hard negative text mining still matches their final performance. In this paper, NegCLIP refers to our replicated version which doesn't involve hard image negative mining.} The resulting model is called NegCLIP. They also fine-tuned CLIP on COCO normally without any additional hard negatives, resulting in the model CLIP\_FT. Table \ref{tab:hard-test-acc} shows the VG Attribution accuracies of these three models, and NegCLIP clearly outperforms CLIP\_FT and CLIP. Since these models are obtained in very similar ways, we decided to use our framework to understand where this improvement comes from.

\begin{figure*}[tbp]
\centering
\subfigure[CLIP]{\label{fig:clip-a}\includegraphics[width=55mm]{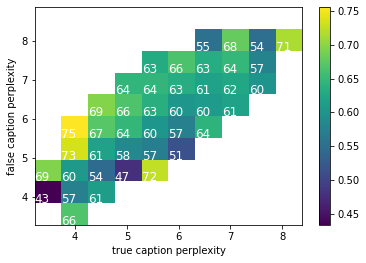}}
\subfigure[CLIP\_FT]{\label{fig:clip-ft-a}\includegraphics[width=55mm]{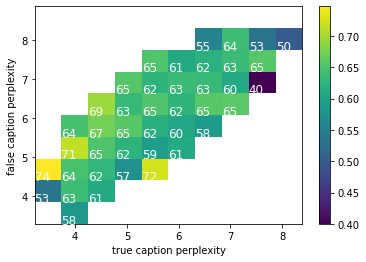}}
\subfigure[NegCLIP]{\label{fig:clip-neg-a}\includegraphics[width=55mm]{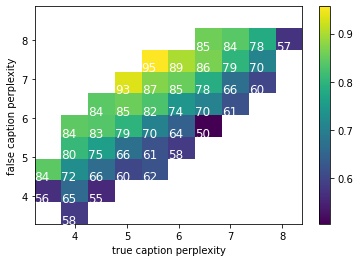}}
\caption{Prediction Accuracies of CLIP variants based on language prior. The $x$ and $y$ axis are the perplexities (computed by AVG\_LM) of true and false captions, separately, and the number in each square shows the prediction accuracy (in percentile) of the vision-language model on data belonging to that particular region.}
\label{fig:align-prior-2d}
\end{figure*}

\subsection{Prediction Accuracies of CLIP variants based on language prior}

Similar to Section \ref{sec:LM_performance}, we use the perplexity computed by AVG\_LM to represent the possibility of a caption, and compute fine-grained model performance for input data that have similar perplexities for the caption candidates. Specifically, for VG Attribution, we divide the range of true/false caption perplexities evenly into 10 intervals\footnote{Note that the true caption has a slightly smaller average perplexity so the x-axis and y-axis are not perfectly aligned.}, and this divides the input data into 100 blocks. We only compute the accuracy when there are at least 10 samples in a block.

Figure \ref{fig:align-prior-2d} shows the prediction accuracies of pre-trained CLIP, CLIP\_FT, and NegCLIP based on linguistic priors. The smaller the true caption perplexity is, or the larger the false caption perplexity is, the easier it is for language models to select the true caption between the two candidates. In other words, the top-left parts are the easiest for language models, and the bottom-right parts are the hardest.

From Figure \ref{fig:align-prior-2d} we could notice that the prediction accuracies of both CLIP and CLIP\_FT are relatively stable for different blocks, meaning their compositional generalization performance is mostly independent from the perplexities of the caption candidates. However, for NegCLIP, the prediction accuracies for upper-left blocks are much higher than those of bottom-right ones. Therefore, the performance of NegCLIP has a strong correlation with the linguistic prior. Furthermore, despite the improvement in overall accuracy, the performance of NegCLIP for bottom-right blocks are similar to CLIP and CLIP\_FT. Hence we hypothesize that NegCLIP mostly learns linguistic priors during its fine-tuning process.

\subsection{Hard Test Accuracy and Linguistic Gap}


To further quantify the alignment of these models with linguistic priors, we propose a new metric ``hard test accuracy'' to measure the dependency of the benchmark performance of models on linguistic priors of the captions.

To eliminate the effect of linguistic priors, we only selected the samples which are hard for the language models to classify, and evaluate the performance of vision-language models on these hard instances. For ARO datasets, we define ``hard instances'' to be the data where AVG\_LM made mistakes, i.e., where the true caption has a higher perplexity than false ones. The number of hard instances is about 15\% of the original dataset. We define ``hard test accuracy'' as the accuracy of a model on these hard instances, and define ``linguistic gap'' as the difference between the overall accuracy and the hard test accuracy.

The linguistic gap can reflect the dependency of the model's performance on linguistic priors. A model whose prediction is independent of the perplexities of the candidate captions should have a constant accuracy over hard and easy samples, yielding zero linguistic gap. A model relying on linguistic priors will be more accurate when the true caption has a lower perplexity than the false captions, so it will have a positive linguistic gap. Therefore, for models with a fixed overall accuracy, a larger linguistic gap implies a stronger dependency on linguistic priors.

Table \ref{tab:hard-test-acc} shows the overall and hard test accuracies of CLIP variants on VG Attribution. Similar to our intuition from Figure \ref{fig:align-prior-2d}, the linguistic gap for CLIP and CLIP\_FT is significantly smaller than that of NegCLIP. We also notice that the hard test accuracy of NegCLIP is almost the same as CLIP\_FT, so the effect of the extra negative captions only helps the model learn linguistic priors and does not help with the interplay between image and text.

\begin{table}[htbp]
\caption{Accuracies and Linguistic Gap on VG Attribution}
\label{tab:hard-test-acc}
\vskip 0.15in
\begin{center}
\begin{small}
\begin{sc}
\begin{tabular}{lccc}
\toprule
 & Total Acc & Hard Acc & Ling. Gap \\
\midrule
CLIP &  61.35 & 57.92 & 3.43\\
CLIP\_FT & 63.17 & 60.43 & 2.74\\
NegCLIP &  74.90 & 61.66 & 13.24\\
\midrule
BLIP ITC  & 80.00 & 76.63 & 3.37\\
BLIP ITM  & 88.56 & 88.08 & 0.48\\
FLAVA Contr  & 60.37 & 58.91 & 1.46\\
FLAVA ITM &  68.83 & 71.10 & -2.27 \\
\bottomrule
\end{tabular}
\end{sc}
\end{small}
\end{center}
\vskip -0.1in
\end{table}


\subsection{Hard Test Accuracy on the VG Attribution Dataset}

Using this framework, we also evaluate the hard test accuracies for commonly-used vision-language models, including different heads of BLIP and FLAVA, as shown in Table \ref{tab:hard-test-acc}. Although different models and different heads vary in their overall prediction accuracies, these models do not have a large linguistic gap, so their prediction on VG Attribution is relatively independent of the linguistic prior.

This framework can be applied on other ARO datasets as well. The conclusions are similar to VG Attribution, and the detailed results are in Appendix \ref{sec:other-ARO}.

\section{Conclusion and Future Work}
\label{sec:conclusion}
In this paper, we studied the multi-modal compositional generalization ability of vision-language models. We showed that language models with similar sizes could achieve better performances than vision-language models on vision-language compositional generalization benchmarks without even taking the images as inputs, demonstrating the important role of linguistic priors in vision-language compositional generalization. We then decomposed the model performance on these benchmarks into a uni-modal part and a multi-modal part, and proposed a new metric to quantify these two parts. 

For future directions, we would like to apply our framework to more models, benchmarks, and modalities. It will also be interesting to explore better ways of generating hard negatives for these models. Finally, it is more important to dig deeper into the reasons why multi-modal networks have this compositional generalization behavior. The reasons may lie in the shared embedding space for different modalities.


\bibliography{example_paper}
\bibliographystyle{icml2023}

\newpage
\appendix
\onecolumn

\section{Hard Test Accuracy on other ARO datasets}
\label{sec:other-ARO}

We also apply the framework of hard test accuracy on the other ARO datasets. However, since language models perform almost perfectly on the Ordering datasets, the number of hard instances is very small (400+ for COCO ordering and less than 100 for Flickr Ordering). As for VG Relation, when evaluating hard test accuracy, we only keep the relationships(``hard relationships'') that contain at least 10 hard instances, which reduces the number of relationships by 80\%. Therefore, for VG Relation, we compute both the prediction accuracy on all data in hard relationships (hard total accuracy) and on hard instances in hard relationships (hard test accuracy). The results are shown in Table \ref{tab:hard-test-acc-RO}. The results for Flickr Ordering are very close to COCO Ordering and we omit them due to space constraints.

Similar to VG Attribution, both CLIP and CLIP\_FT have much smaller linguistic gaps than NegCLIP, and NegCLIP even has the worst hard test accuracy, meaning it aligns so much with the linguistic prior that the prediction power on hard instances got reduced.

\begin{table}[htbp]
\caption{Overall and Hard Accuracies on COCO Ordering and VG Relation, format: ``overall accuracy / hard test accuracy'' for COCO Ordering, ``overall accuracy / hard overall accuracy / hard test accuracy'' for VG Relation.}
\label{tab:hard-test-acc-RO}
\begin{center}
\begin{small}
\begin{sc}
\begin{tabular}{lccc}
\toprule
 & COCO\_Order & VG Relation \\
\midrule
CLIP & 47.73 / 43.56   & 59.79 / 53.52 / 55.90\\
CLIP\_FT & 31.18 / 27.10   & 60.01 / 53.11 / 50.48\\
NegCLIP & 91.74 / 71.56  & 80.50 / 60.94 / 48.62 \\
\midrule
BLIP ITC  & 13.21 / 17.51 & 41.59 / 52.02 / 58.31 \\
BLIP ITM  & 20.83 / 23.95 & 53.35 / 62.26 / 64.80 \\
FLAVA Con  & 7.03 / 4.79 & 29.13 / 38.17 / 36.00 \\
FLAVA ITM & 4.81 / 4.79 & 23.71 / 37.20 / 33.91 \\
\bottomrule
\end{tabular}
\end{sc}
\end{small}
\end{center}
\end{table}

\section{More Experimental Details}

\textbf{Image-Text Matching Score Computation}: For models that output embedding vectors, we compute the inner product of the normalized image and text embeddings and use this number as the matching score. For models that directly compute a score for image-text matching, we use this score directly. After getting the scores, for each datum in the ARO datasets, since the image is fixed, we predict the text with the highest matching score as the true caption and compute the prediction accuracy.

\textbf{Model Details}: We obtain CLIP \cite{radford2021learning}, BLIP \cite{li2022blip}, and FLAVA \cite{singh2022flava} models using the same way as in \cite{yuksekgonul2022and}, and download all language models from huggingface.



\end{document}